\pdfoutput=1

\documentclass[11pt]{article}	

\usepackage{emnlp2021}

\usepackage{times}
\usepackage{latexsym}

\usepackage[T1]{fontenc}

\usepackage[utf8]{inputenc}

\usepackage{microtype}

\usepackage{svg}
\usepackage{kotex}
\usepackage{booktabs}
\usepackage{colortbl}
\usepackage{tikzsymbols}
\usepackage{booktabs}
\usepackage{amsmath}
\usepackage{amsfonts}
\usepackage{multirow}
\usepackage{makecell}
\usepackage{graphicx}
\usepackage{paracol}
\usepackage{subfig}
\usepackage{graphicx}
\usepackage{adjustbox}
\usepackage{kotex}
\usepackage{float}

\definecolor{orange}{rgb}{1,0.5,0}
\definecolor{goldenpoppy}{rgb}{0.6, 0.4, 0.08}
\definecolor{ku}{rgb}{0.0, 0.26, 0.15}
\definecolor{jhu}{rgb}{1.0, 0.66, 0.07}
\definecolor{deepfuchsia}{rgb}{0.76, 0.33, 0.76}

\makeatletter
\newcommand{\thickhline}{%
    \noalign {\ifnum 0=`}\fi \hrule height 1pt
    \futurelet \reserved@a \@xhline
}
\makeatother

\usepackage{comment}
\usepackage{todonotes}

\newcommand*{\img}[1]{%
    \raisebox{-.3\baselineskip}{%
        \includegraphics[
        height=\baselineskip,
        width=\baselineskip,
        keepaspectratio,
        ]{#1}%
    }%
}

\title{Who speaks like a style of Vitamin \img{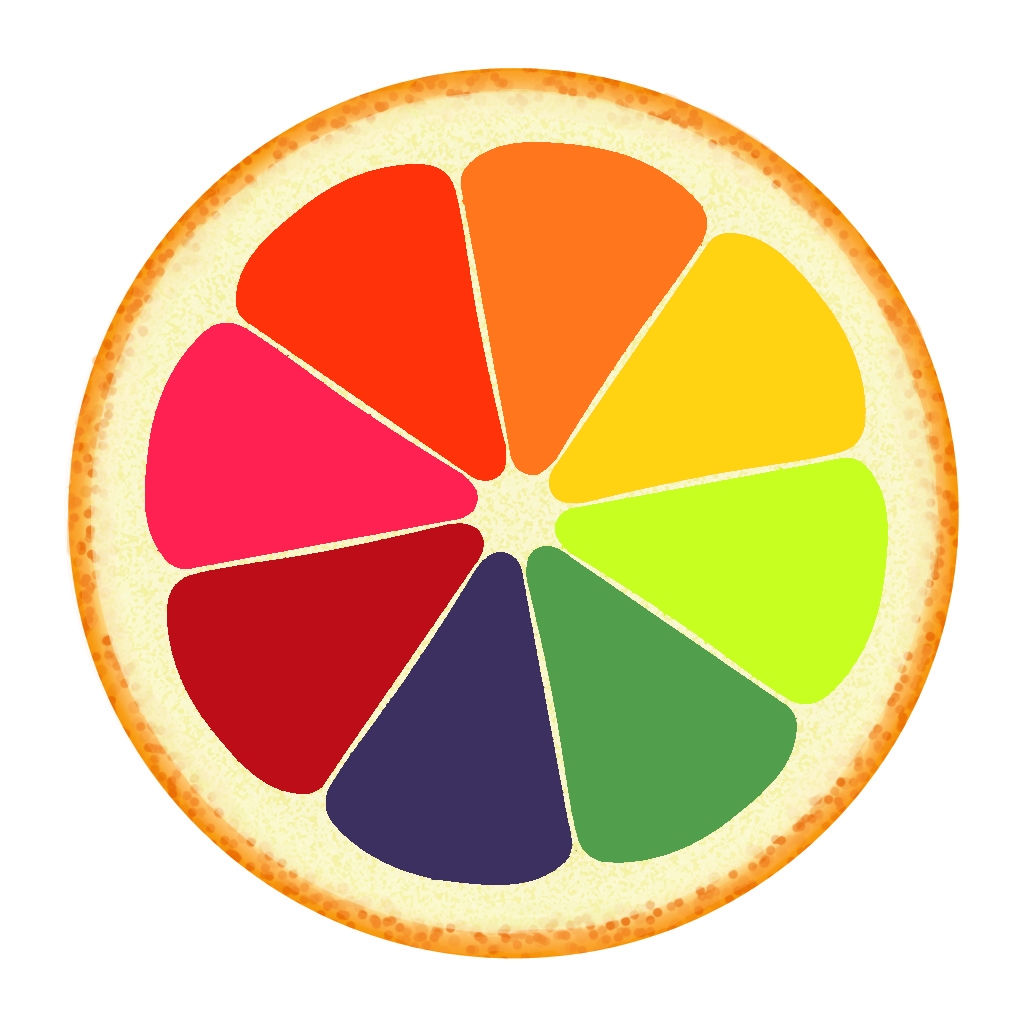}\thanks{\hspace*{1em}We define the speaker styles into Vitamin A,B, and C.} : Towards Syntax-Aware Dialogue Summarization using Multi-task Learning}

\author{Seolhwa Lee$^1$, Kisu Yang$^2$, Chanjun Park$^1$, Jo\~ao Sedoc$^3$, Heuiseok Lim$^1$\\
     $^1$Korea University, South Korea \\
     $^2$VAIV Corp, South Korea \\
     $^3$New York University, United States \\
  \texttt{\{whiteldark, bcj1210, limhseok\}@korea.ac.kr} \\
  \texttt{\{ksyang\}@vaiv.kr} \\
  \texttt{\{jsedoc\}@stern.nyu.edu}}

\begin{document}
\maketitle
\begin{abstract}
Abstractive dialogue summarization is a challenging task for several reasons. First, most of the important pieces of information in a conversation are scattered across utterances through multi-party interactions with different textual styles. Second, dialogues are often informal structures, wherein different individuals express personal perspectives, unlike text summarization, tasks that usually target formal documents such as news articles. 
To address these issues, we focused on the association between utterances from individual speakers and unique syntactic structures. Speakers have unique textual styles that can contain linguistic information, such as voiceprint. Therefore, we constructed a syntax-aware model by leveraging linguistic information ({\em{i.e.,}} POS tagging), which alleviates the above issues by inherently distinguishing sentences uttered from individual speakers.
We employed multi-task learning of both syntax-aware information and dialogue summarization. To the best of our knowledge, our approach is the first method to apply multi-task learning to the dialogue summarization task. Experiments on a SAMSum corpus (a large-scale dialogue summarization corpus) demonstrated that our method improved upon the vanilla model. We further analyze the costs and benefits of our approach relative to baseline models.
\end{abstract}

\section{Introduction}
Humans interactively converse every day with other humans, and more recently with machines. Humans constantly converse with each other, online or offline, for example, when attending meetings or using customer service. Therefore, textual dialogues are an essential component of the interactions between users and user agents.

This abundance of personal and public conversations represents a valuable source of information, but analyzing such an immense amount of data to meet specific information needs can lead to information overload problems~\cite{jones2004information}.

\begin{figure}[t!]
\begin{center}
\includegraphics[width=1\linewidth]{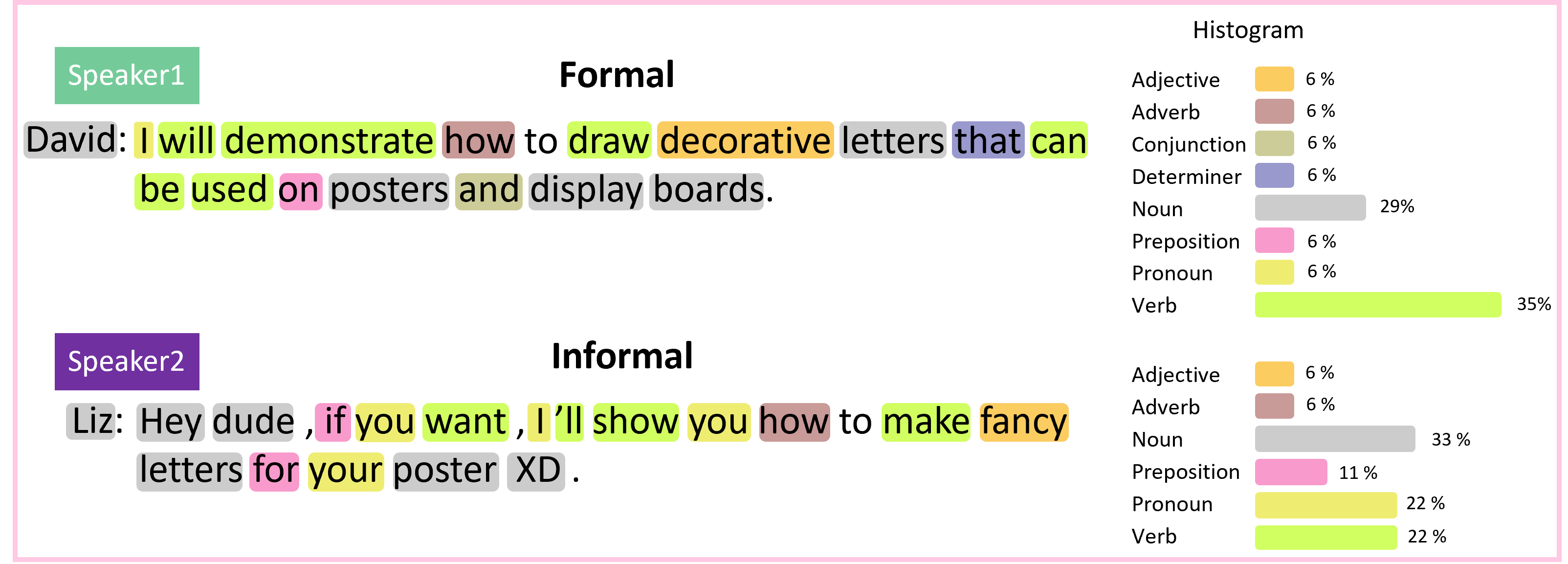}
\end{center}
\caption{Example utterance of formal and informal sentences of the same meaning from different speakers, with the different parts-of-speech labeled. The histogram shows the different individual textual styles.} 
\label{fig:pos_example}
\end{figure} 
Recently, dialogue summarization has emerged as a means to resolve this issue. Dialogue summarization is the task of distilling the highlights of an instance of a dialogue and rewriting them into an abridged version. This task is closely associated with abstractive text summarization~\cite{shi2021neural,gehrmann-etal-2018-bottom,nallapati2016abstractive}.

However, most of the existing investigation efforts on abstractive text summarization have been concentrated on single-speaker documents, such as news articles~\cite{paulus2017deep,see2017get} and scientific documents~\cite{cachola-etal-2020-tldr, erera-etal-2019-summarization}.  Research has largely focused on the composition of strictly formatted paragraphs ({\em{e.g.,}} introduction-method-conclusion) or is simply dependent on locational biases ({\em{i.e.,}} the tendency towards a particular location such as the lead or tail of the text)~\cite{kim2019abstractive}. 

Unlike these strictly structured formats, the format of dialogues between multiple interlocutors is often informal. Dialogues are represented through a variety of utterances, including the expression of personal perspectives, opinions, markers of certainty and doubt, speaker interruptions~\cite{precht2008sex,sacks1978simplest}, colloquial representations, and even the use of emojis in a textual way, and important information is scattered throughout  dialogue, not necessarily restricted to predictable areas. These features lead to a focus on informative utterances. Figure~\ref{fig:pos_example} presents the representation of different speaker styles for informal and formal utterances that we attempt to address in this paper.

Thus, the critical challenges of dialogue summarization task are: 1) \textit{\textbf{Multiple speakers and the different textual styles;}} the essential pieces of information in a conversation are scattered across the utterances of interlocutors through their different textual styles~\cite{koppel2009computational}. 2) \textit{\textbf{Informal structure;}} dialogues consist of an informal structure, including slang and colloquial language, free of the locational biases found in formal structures~\cite{chen2020multi}.

To address these challenges, we investigated the relationship between textual styles and representative attributes of utterances.  \citet{kubler2010adding} proposed that the types ({\em{e.g.,}} intent or role of a speaker) of sentences from speakers are associated with different syntactic structures ({\em{i.e.,}} linguistic information), such as part-of-speech (POS) tagging. This is derived from the fact that different speaker roles are characterized by different syntactic structures.

Research in dialogue summarization is benefited from research in related fields, such as speaker recognition. Speaker recognition is the process of identity recognition, and specifically uses identity information ({\em{i.e.,}} voiceprint) from the human voice. The speaker's speech signal is considered as the hidden features that can be used to distinguish the identity of that individual and is commonly used in the speaker recognition research field~\cite{guo2021speaker,liu2018gmm}. In essence, the uttered text has a unique representation from each speaker, like a voiceprint. Based on this prior research, we began our study with the assumption that because syntactic structures tend to be associated with a representative of a sentence uttered from speakers, these structures would help distinguish the different styles of utterances. This assumption was also maintained in previous research. \citet{zhu2020hierarchical} proposed a hierarchical structure to handle the transcripts of long meetings and adopted a role vector to represent the individual speakers in conversation summarizations.

Inspired by the previous works, we propose a novel abstractive dialogue summarization model for use in a daily conversation setting, which is characterized by an informal style of text, including emoticons and abbreviations of chat terms. Furthermore, we explore the locational biases in dialogue structures. Although dialogues show independent locational biases different from that of formal documentation, we evaluated different simple baselines based on locational biases motivated by~\citet{gliwa2019samsum,kim2019abstractive} (see details in Section~\ref{subsec:implementation}). The main contributions of this paper are fourfold. 

\begin{itemize}
    \item First, we propose a novel approach for the abstractive dialogue summarization task. Specifically, the multi-task learning model is proposed to learn abstractive dialogue summarization and perform sequence labeling tasks simultaneously, to reflect the syntactic features on the dialogue summarization model. 
    \item Second, to the best of our knowledge, this is the first study to perform multi-task learning on the dialogue summarization task using the SAMSum corpus and, specifically, to integrate these tasks using linguistic information.
    \item Third, we propose a novel input type training method, rather than using the traditional method of truncating the input, to investigate locational biases. 
    \item Finally, the proposed method outperformed the base models for all ROUGE scores~\cite{lin2004rouge}.
\end{itemize}

\section{Proposed Method}
\subsection{Why Part-of-speech-tagging?}
Part-of-speech-tagging is a valuable resource for analysis in the syntax-aware approach. \citet{arifin2018sentence} conducted multi-document summarization to find representative sentences, not only by sentence distribution to select the most important sentences but also by how informative a term is in a sentence. They used part-of-speech (POS) tagging information to resolve this and obtain improved performance. This approach is characterized by its use of grammatical information, which is carried by POS labels, and the presence or absence of informative content in a sentence. 

We further considered that incorporating the linguistic information from POS tagging could help alleviate structure/context ({\em{i.e.,}} formal and informal) dependency issues for text summarization, and also in dialogue summarization. Simultaneously learning the syntax-aware approach using linguistic information and language generation allows the sharing of grammatical information that constrains next word generation.

Also, it is possible to deal with the first challenge by applying syntax-awareness to the entirety of the utterances from the dialogue because this will recognize the linguistic information from the speakers and also intrinsically represent the textual styles. Therefore, the model obtains the built-in ability to distinguish text styles.

\subsection{Problem Formulation}
We formalize the problem of dialogue summarization as follows. The input consists of dialogues $\chi$ and dialogue speakers $S$. Assume there are $d$ dialogues in total. The dialogues are $\chi = \{X_1,...,X_d\}$. Each dialogue consists of multiple turns, where each turn is the utterance of a speaker. Therefore, $X_i = \{(s_1, u_1), (s_2, u_2), ..., (s_{L_i}, u_{L_i})\}$, where $s_j \in S$, $1\leq j \leq L_i$, is a speaker and $u_j = (w_1,...,w_{l_j})$ is the tokenized utterance from $s_j$. The human-annotated summary for dialogue $X_i$, denoted by $Y_i$, is also a sequence of tokens. In the end, the aim of the task is to generate a dialogue summary $\hat{Y}=(\hat{y}_1, ..., \hat{y}_l)$ given the dialogues $X=\{(s_1, u_1), (s_2, u_2), ..., (s_m, u_m)\}$ and the reference summaries $Y=(y_1, ..., y_k)$. 

The purpose of the sequence labeling task is to predict the sequence label $Y_{POS}=(pos_1, ..., pos_n)$, where $pos_i=\{tag_1,...,tag_{l_j}\}$, and $|pos|$ is the number of tags $tag$ in an utterance.

To summarize, the final goal of dialogue summarization is to maximize the conditional probability of the dialogue summary $Y$, given dialogues $X$ and model parameters $\theta: P(Y|X;\theta)$. 

\subsection{Preprocessing for Syntax-Aware SAMSum} 
We automatically annotated sequence labels for all the utterances as these are not included in the SAMSum corpus. We used these labels for training syntax-aware information using the steps below.
\paragraph{Tokenization for data labeling} To better recognize syntax-aware information, we used Twokenizer\footnote{\url{https://github.com/myleott/ark-twokenize-py} Note that Twokenizer is used for data labeling not for model training.} \cite{owoputi2013improved} before annotating the sequence labeling. The Twokenizer was revised for tweet text to conduct part-of-speech (POS) tagging. Tweet text consists of online conversational text that also contains many nonstandard lexical items and syntactic patterns, such as emojis and emoticons. Also, the daily chat includes those like tweet text.
Emoticons ({\em{e.g.,}} :), XD) refer to the generation of a face or icon using traditional alphabetic or punctuation symbols, whereas emojis ({\em{e.g.,}} \cChangey{2}, \cChangey{-1}) refer to when small pictures used as symbols. The Twokenizer accurately recognizes emoticons as, for example, ``:)'' not as ``: ).''
\paragraph{Part-of-speech tagging} We obtained tokenized utterances using the above method. This process improves the model's ability to recognize each token to use the POS tagger. For the sequence labeling, we adopted the CMU-Twitter-POS-tagger\footnote{\url{http://www.cs.cmu.edu/~ark/TweetNLP/}}~\cite{owoputi2013improved,gimpel2010part}, which addresses the problem of POS tagging for English data from the popular micro-blogging service Twitter.

\begin{figure*}[h!]
\begin{center}
\includegraphics[width=0.9\linewidth]{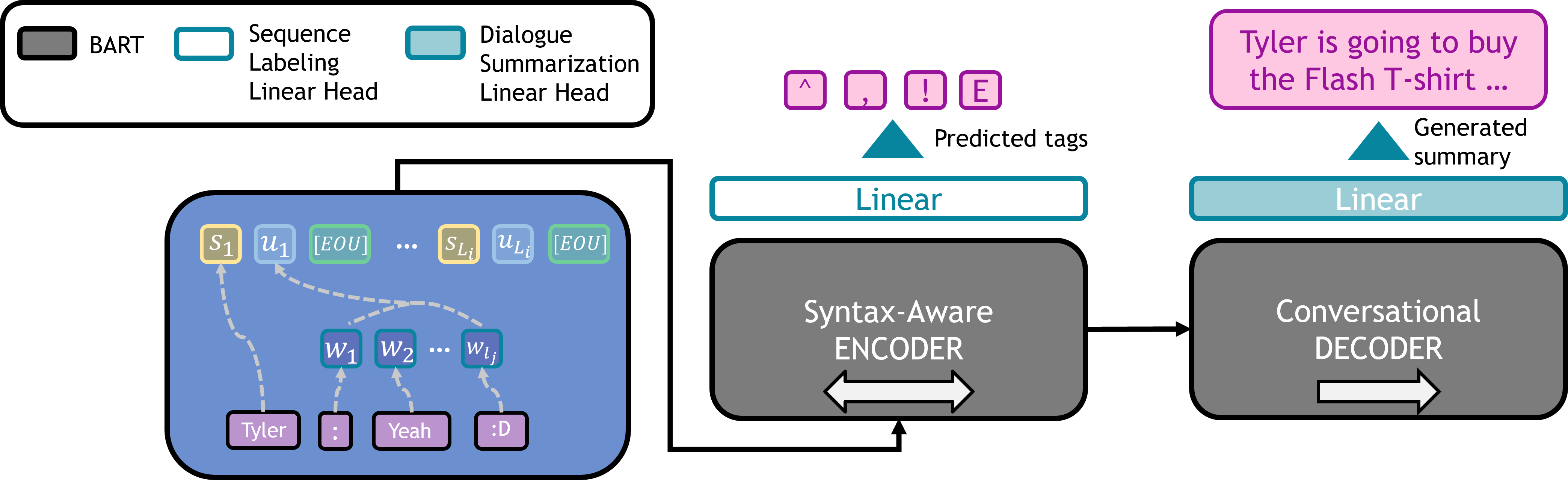}
\end{center}
\caption{Overview of the model architecture. The syntax-aware encoder with a task-specific linear head learns the sequence labeling task given the dialogue utterances in a bidirectional encoder setting from the BART encoder. The conversation decoder ({\em{i.e.,}} autoregressive decoder from the BART decoder) learns the dialogue summarization task through the linear head.} 
\label{fig:overall}
\end{figure*} 

\subsection{Model Overview}
Regarding the multi-task learning for BART backbone, we address two different tasks simultaneously: token classification ({\em{i.e.,}} sequence labeling) and language modeling ({\em{i.e.,}} generation). BART consists of a bidirectional encoder and an autoregressive decoder. Therefore, we conducted the token classification task in the encoder ({\em{i.e.,}} syntax-aware encoder) and the language model task in the decoder ({\em{i.e.,}} conversational decoder). As illustrated in Figure~\ref{fig:overall}, task-specific linear heads were trained through multi-task learning, which performs the main task as a dialogue summarization task and the POS sequence labeling task as an auxiliary task.

\subsection{Syntax-Aware Encoder} \label{sec:syntax-aware-encoder}
We sought to address the application of syntax-aware information to a dialogue summarization model through the sharable encoder.

Each utterance $u_i$ was composed of a special [EOU] token, as was done in previous work~\cite{gliwa2019samsum}, considering each utterance separately. In general, the input sequence,
\begin{equation}
\mathrm{{X_i}^{'}}=\{s_1;u_1;[EOU] ... s_{L_i};u_{L_i};[EOU]\},
\label{eq:equ-1}
\end{equation}
is fed into the bottom encoder $\mathrm{\bf E}$ of BART. Given the hidden outputs of the encoder's last layer $\{{h_1}^{L}, ..., {h_n}^{L}\}$, the output layer for the sequence labeling task was a linear classifier $f : \mathbb{R}^{d} \to \mathcal{Y}$, where $\mathbb{R}^{d}$ denotes the dimension of the hidden layer and $\mathcal{Y}$ is a $(k-1)$ simplex, where $k$ is the number of POS tags.
\begin{equation}
\begin{aligned}
\{{h_1}^{L}, ..., {h_n}^{L}\} = \mathrm{\bf E}(\{\mathrm{{X^{'}_1}, ..., {X^{'}_n}\}}),
\label{eq:equ-2}
\end{aligned}
\end{equation}
, where $L$ is the last layer. The probability that the word $w_{l_j}$ aligns with the $m$-th POS tag is computed using softmax:
\begin{equation}
\begin{aligned}
P(m|w_{l_j}) = \mathrm{Softmax}(W_m{h_n}^{L}),
\end{aligned}
\end{equation}
, where $W_m$ is a parameter to be learned.

\subsection{Conversational Decoder}
To integrate the syntax-aware encoder with the decoder, the dialogue summarization model consists of combining the shared syntax-aware encoder using the Equation~\ref{eq:equ-2} and the conversational decoder.
\paragraph{Shared Syntax-Aware Encoder} We used the same encoder over all-around layers from the Section~\ref{sec:syntax-aware-encoder} to apply the syntax-aware information to the dialogue summarization model. Both encoders from each task were shared.

The syntactic information could provide different conversational aspects for the models to learn and further determine which set of utterances deserve more attention to generate better dialogue summaries. 

The input to the decoder included $l-1$ previously generated tokens $\{t_1, ..., t_{l-1}\}$. We fed the tokens to the conversation decoder $\mathrm{\bf D}$, and the $l$-th token ($\hat{t}_l$) was predicted as follows:
\begin{equation}
\begin{aligned}
\tilde{\mathrm{\bf E}} = \mathrm{\bf E}(\{\mathrm{{X^{'}_1}, ..., {X^{'}_n}\}}),
\label{eq:equ-3}
\end{aligned}
\end{equation}
\begin{equation}
\begin{aligned}
\{\hat{y}_1, ..., \hat{y}_{l-1}\} = \mathrm{\bf D}(\{{t_1}, ..., {t_{l-1}}\}, \tilde{\mathrm{\bf E}}),
\label{eq:equ-4}
\end{aligned}
\end{equation}
\begin{equation}
\begin{aligned}
P(\hat{t}_l|t_{<l}, X^{'}) = \mathrm{Softmax}(W\hat{y}_{l-1}),
\label{eq:equ-5}
\end{aligned}
\end{equation}
, where $W$ is a parameter to be learned.

\subsection{Syntax-Aware Multi-task Learning}
We trained the two tasks jointly using multi-task learning. We considered the dialogue summarization task as the main task and the sequence labeling task as an auxiliary task.
\paragraph{Joint training} During training, the two self-learning objectives were combined with the cross-entropy loss for each task, and we sought to minimize the loss as follows:
\begin{equation}
\begin{aligned}
\mathcal{L_{DS}} = -\sum\mathrm{log}P({y_l}|\hat{t}_{<l}, X^{'}),
\end{aligned}
\end{equation}
\begin{equation}
\begin{aligned}
\mathcal{L_{POS}} = -\sum\mathrm{log}P(pos_{l_j}|m),
\end{aligned}
\end{equation}
Thus, the final loss $\mathcal{L}$ of our model is:
\begin{equation}
\begin{aligned}
\mathcal{L} = \lambda\mathcal{L_{POS}} + (1-\lambda)\mathcal{L_{DS}}.
\end{aligned}
\end{equation}
, where $\mathcal{L_{DS}}$ and $\mathcal{L_{POS}}$ are the loss of the dialogue summarization model and the sequence labeling model, respectively, and $\lambda$ denotes the parameter of strength in each task. 

Finally, the model can activate linguistic information to enhance its ability to distinguish a speaker's utterance style.

\subsection{Speaker Styles of Utterance for Ad-hoc Analysis}
\label{sec:speaker_style}
In order to represent the uttering styles of speakers, we considered a list of POS tags extracted from utterances of each speaker as a style of the speaker as Equation \ref{eq:style_is_a_set_of_tags} where $i$ denotes the index of speaker, and $T$ is the number of tags. It has the same form of a document made up of words.
\begin{equation}
\label{eq:style_is_a_set_of_tags}
{style_{i}}=\{tag_1, ..., tag_T\},
\end{equation}

With the style documents, we conducted tf-idf, a commonly used method to weight the importance of each term in a document. The tf-idf formula is as follows:
\begin{equation}
\label{eq:tfidf}
W_{ij}=tf_{ij}*idf_{ij}=tf*log(\frac{n}{1+df_{ij}}),
\end{equation}
In Equation \ref{eq:tfidf}, $tf_{ij}$ represents a term frequency of the $j$-th tag in the $i$-th speaker style, and $df_{ij}$ denotes the number of $i$-th speaker styles in which the $j$-th tag appears, and $n$ is the total number of speaker styles. Then we employed K-means clustering for grouping the speaker styles. Consequently, this ad-hoc analysis was used to show whether our proposed strategies have been worked as intended in the trained models.


\section{Experimental Setup}
\subsection{Dataset and Baselines}
We trained and evaluated our model on a large-scale dialogue summary dataset SAMSum~\cite{gliwa2019samsum}. SAMSum is the first daily chat corpus for use in dialogue summarization and truly informal conversation. The subject of each conversation is open domain, and the conversation type is informal.
The details on data statistics in SAMSum corpus are shown below, including in Table~\ref{tab:data_statistics}.

\begin{table}[tbh!]
\centering
\scalebox{0.95}{
\resizebox{\columnwidth}{!}{
\begin{tabular}{rl|cc|cc|cc} 
\hline
\multicolumn{2}{r|}{\multirow{2}{*}{\textbf{\# Conv}}} & \multicolumn{2}{c|}{\textbf{S.L}} & \multicolumn{2}{c|}{\textbf{\# Speakers}} & \multicolumn{2}{c}{\textbf{\# Turns}} \\
&  & Mean & Range & Mean & Range & Mean & Range  \\ \hline
Train & 14732 & 23.44 & [2, 73] & 2.40 & [1, 14] & 11.17 & [1, 46]      \\
Dev & 818 & 23.42   & [4, 68] & 2.39 & [2, 12]  & 10.83  & [3, 30]  \\
Test & 819 & 23.12   & [4, 71] & 2.36 & [2, 11] & 11.25  & [3, 30]      \\ \hline 
\end{tabular}}}
\caption{Data statistics of the SAMSum corpus. S.L denotes summary length. Range indicates the minimum and maximum values.} \label{tab:data_statistics}
\end{table}

\paragraph{Data Statistics} To better understand the characteristic of this corpus, we explored the density of the number of utterances. 

As shown in Figure~\ref{fig:utter-statistic} (see Appendix~\ref{appendix:density}), the density of the number of utterances is consistently below ten. This result indicates that the utterances are generally fewer than ten in the training set. This result allows us to choose the locational biases, like input sequence types such as LEAD-3~\cite{see2017get}, which takes the three leading sentences of the source text as the summary (we discuss this in Section~\ref{subsec:implementation}).

\paragraph{Baselines} We evaluated the model's performance with the following summarization models, based upon previous works~\cite{gliwa2019samsum}.

\begin{itemize}
  \item \textbf{Pointer Generator}~\cite{see2017get} This model followed~\citet{gliwa2019samsum}, wherein separators are added between each utterance and utilized as input for the pointer generator model.
  \item \textbf{DynamicConv + GPT-2}~\cite{wu2019pay} Based on ~\citet{gliwa2019samsum}, this model uses GPT-2 to initialize token embeddings~\cite{radford2019language}.
  \item \textbf{Fast Abs RL Enhanced}~\cite{chen2018fast} adopts a hybrid method that selects salient sentences and then paraphrases them as abstractive sentences through sentence-level policy gradient methods. 
  \item \textbf{BART}~\cite{lewis-etal-2020-bart} 
  We used BART as the vanilla in the following setting and added a separator in each utterance. The default parameter setting was \texttt{BART-base}\footnote{\url{https://huggingface.co/ transformers/model_doc/bart.html}}. Additionally, we fed the input type with the LONGEST-10 settings, as described in Section~\ref{subsec:implementation}.
\end{itemize}

\subsection{Evaluation Metrics}
We utilized different evaluation metrics, including several recently introduced methods used in text summarization and generation tasks.

\paragraph{ROUGE-N\footnote{\url{https://github.com/pltrdy/rouge} Note that different packages may generate different ROUGE scores.}}~\cite{lin2004rouge} mostly used evaluation metrics for the text summarization task. We calculated ROUGE-1, ROUGE-2, and ROUGE-L. 
\paragraph{BertScore\footnote{\url{https://pypi.org/project/bert-score/}}}~\cite{zhang2019bertscore} calculates the aligning similarity scores between the generated and reference summaries on a token level using BERT. 

\subsection{Implementation Details}\label{subsec:implementation}
We tested different input type modes in our proposed model. The traditional method for handling long sequences in a pretrained language model is to truncate the sequence in an uncompleted format, not in the true utterance format, owing to limitations in system memory. To alleviate this issue, we propose a novel input type method to retain the utterance format.
Inspired by previous work~\cite{gliwa2019samsum,see2017get}, we defined the input types as LEAD-n, MIDDLE-n, and LONGEST-n. The underlying assumption of these input types is that the locational biases~\cite{kim2019abstractive} contain the essential information at the head ({\em{i.e.,}} the beginning of the lead) or middle of lengthy conversations. Additionally, this method preserves the entire utterance sequences without breaking up sentence information. To support the above assumptions, we performed data statistics as described in the above section. We trained the model according to the different settings described as follows: \textbf{LEAD-n} - takes $n$ leading utterances of the dialogues, \textbf{MIDDLE-n} - takes $n$ utterances from the middle of the dialogue, and \textbf{LONGEST-n} - takes $n$ utterances from the longest of the dialogue.
We used the \texttt{BART-base} model to initialize the backbone of the encoder/decoder frame and followed the default settings. The learning rate was set to 3e-4. We trained the model for 20 epochs. Also, we set $\lambda$ as 0.1 in the final model. The training was conducted on a single RTX 8000 GPU with 48 GB memory. We trained the model with the Adam optimizer~\cite{kingma2014adam} and an early stop on validation set ROUGE-1. During the inference, the beam size was 4, including for the baseline models.

\section{Main Results}
\subsection{Quantitative Results}
We evaluated the models across the different settings with ROUGE-1, ROUGE-2 and ROUGE-L, and BertScore on the SAMSum test set. The experimental results are shown in Tables~\ref{tab:result-type} and ~\ref{tab:result}. 

\begin{table}[h!]
\centering
\resizebox{\columnwidth}{!}{
\begin{tabular}{l|c|l|cccc}
\hline
\textbf{Type} & $\lambda$ & \textbf{n} & \textbf{ROUGE-1} & \textbf{ROUGE-2} & \textbf{ROUGE-L} & \textbf{BertScore} \\ \hline
\textbf{LEAD}& \multirow{3}{*}{0.5} &\multirow{3}{*}{10} &0.409&0.170&0.390& 0.909    \\
\textbf{MIDDLE} &  &&0.403&0.167&0.382&0.908     \\
\textbf{LONGEST} &&&\textbf{0.425}&0.183&0.405&0.909  \\ \hline
\textbf{LEAD}&\multirow{2}{*}{0.5}& \multirow{2}{*}{20}&\textbf{0.425}&\textbf{0.188}&\textbf{0.409}& \textbf{0.910}    \\
\textbf{MIDDLE}&&& 0.414 &0.181&0.404&0.908      \\\hline \hline
\textbf{LEAD}& \multirow{3}{*}{0.1} &\multirow{3}{*}{10} &0.426&0.188&0.414& 0.910    \\
\textbf{MIDDLE} & &&0.428&0.192&0.414&0.910      \\
\textbf{LONGEST}&&&\textbf{0.431}&\textbf{0.189}&\textbf{0.420}&\textbf{0.910}  \\ \hline
\textbf{LEAD}&\multirow{2}{*}{0.1}& \multirow{2}{*}{20}&0.424&0.187&0.415& 0.909    \\
\textbf{MIDDLE}&&&0.425&0.189&0.416&0.909      \\\hline
\end{tabular}} \caption{Performance comparison according to the different input type settings for training. $\lambda$ and n indicate the strength of the task ability and the number of utterances in dialogue, respectively.} \label{tab:result-type}
\end{table}

\begin{table}[tbh!]
\centering
\scalebox{0.7}{
\begin{tabular}{c|c|c}
\toprule
\textbf{Model}                        & \textbf{Type}      & \textbf{avg \# words} \\ \hline
Ground summary                      & -         & 23.12        \\ \hline
BART                         & LONG-10   & 22.25        \\ \hline
\multirow{3}{*}{\shortstack[c]{Syntax-aware BART\\($\lambda=0.1$)}} & LEAD-10   & 19.95        \\
                             & MIDDLE-10 & 18.25        \\
                             & LONG-10   & 21.95        \\
\bottomrule
\end{tabular}}
\caption{The average number of words of the generated summaries at an inference.}
    \label{tab:quantitative-2}
\end{table}
\paragraph{Internal model verification} In Table~\ref{tab:result-type}, we explore the influence locational biases in dialogue have on performance. The input type settings are depicted according to the different measures, based on the F1 scores of both ROUGE and BertScore. We set $\lambda$ as 0.5 and 0.1 and compared $n$ of 10 and 20 for each setting\footnote{Note that we did not set the LONGEST-20 due to limitations in computing power.}. With $\lambda$ set as 0.1, the LONGEST-10 model showed the best performance across every measure. However, the performance when $\lambda$ was set as 0.5 was lower than when $\lambda$ was 0.1, in general. Although the BertScore showed a subtle difference, it also showed the highest performance in this result. We interpret this result as indicating that lengthy utterances are valuable when generating summaries, and the key topics are located at the length of 10 in a dialogue. 
\paragraph{Comparison of the generated length} We investigated the length of the generative summary, which varies with the input type. As shown in Table~\ref{tab:quantitative-2}, we examined the average of words according to the generated summaries at each inference step. The BART base model ({\em{i.e.,}} BART$\dagger$) used 22.25 words when it generated the summaries. Moreover, the average words generated by our proposed model according to the input type shows that our best performance model ({\em{i.e.,}} Syntax-aware BART (LONG-10)) used only 21.95 words, thus requiring fewer words than the baseline model. This observation reveals that our proposed model often favors generating slightly shorter summaries than does the BART baseline model, which leads to more concise summaries while still capturing the important information. According to input types, input length influences the average number of words.
\paragraph{External model verification} In Table~\ref{tab:result}, we present the ROUGE-1, ROUGE-2, and ROUGE-L scores between our model and other, comparison models. First, our proposed model outperformed the other baselines with respect to F1 for all ROUGE scores. \textit{As hypothesized previously, our experiments demonstrate that the usage of linguistic information is worthwhile to enhance the model performance.} Fast Abs RL Enhanced achieved slightly better scores than Pointer Generator and DynamicConv+GPT2. This indicates that the use of reinforcement learning to first select important sentences is beneficial. The key factor related to the overall lower performance of the baseline models seems to be that the baseline models fundamentally are not based on the language model; however, the DynamicConv model with the GPT-2 embeddings is based on the usage of pretrained embeddings from the language model GPT-2, which is trained on a large corpus.

\begin{table*}[ht!]
\centering
\resizebox{15cm}{!}{%
\begin{tabular}{c|c|ccc|ccc|ccc}
\hline
\multirow{2}{*}{\textbf{Model}}   & \multirow{2}{*}{\textbf{Type}}          & \multicolumn{3}{c|}{\textbf{ ROUGE-1}} & \multicolumn{3}{c|}{\textbf{ ROUGE-2}} & \multicolumn{3}{c}{\textbf{ ROUGE-L}}\\ \cline{3-11}
                                 &{-}   & F       & P       & R       & F       & P       & R       & F       & P       & R       \\ \hline
Pointer Generator~\cite{see2017get}*    &      -          & 0.401   & -       & -       & 0.153   & -       & -       & 0.366   & -       & -       \\
DynamicConv + GPT-2~\cite{wu2019pay}*     &    -      & 0.418   & -       & -       & 0.164   & -       & -       & 0.376   & -       & -       \\
Fast Abs RL Enhanced~\cite{chen2018fast}* & - & 0.420   & -       & -       & 0.181   & -       & -       & 0.392   & -       & -  \\ \hline \hline
\multirow{1}{*}{BART $\dagger$} & LONG-10  &0.426&0.488&0.419&0.188&0.220&0.184&0.419&0.464&0.415 \\
\multirow{1}{*}{Syntax-aware BART \(\dagger\) (\(\lambda\) =0.1)} & LONG-10  &\textbf{0.431}&0.486&0.426&\textbf{0.189}&0.216&0.186&\textbf{0.420}&0.460&0.418 \\\hline
\end{tabular}%
}
\caption{Performance comparison of the proposed method with different models on the test set. * denotes the results from~\cite{chen2020multi}, and $\dagger$ corresponds to our proposed method model, which shows the best performance (LONGEST-10). Note that F, P, and R indicate F1, precision, and recall scores, respectively.}
\label{tab:result}
\end{table*}


\begin{table*}[h!]
    \centering
    \resizebox{16cm}{!}{%
    \begin{tabular}{|l|l|l|} 
        \toprule
        \bf Dialogue 1 & \bf Dialogue 2 & \bf Speaker Style Utterance (abbreviated) \\\hline
        1. \textcolor{ku}{lilly:} sorry, I'm gonna be late & 
        1. \textcolor{ku}{randolph:} honey & \\ 
        2. \textcolor{ku}{lilly:} don't wait for me and order the food &
        2. \textcolor{ku}{randolph:} are you still in the pharmacy? &  \\
        3. \textcolor{jhu}{gabriel:} no problem, shall we also order & 
        3. \textcolor{jhu}{maya:} yes & (1) Stlye A vs B\\
        something for you? & 
        4. \textcolor{ku}{randolph:} buy me some earplugs please &  ...Robert: ...The Swedes didn't even bother to \textcolor{blue}{find out}...\\
        4. \textcolor{jhu}{gabriel:} so that you get it as soon as you get & 
        5. \textcolor{jhu}{maya:} how many pairs? & they started \textcolor{blue}{laying them off}... (B) \\
        to us? & 
        6. \textcolor{ku}{randolph:} 4 or 5 packs & Cynthia: ...i'd like us to go to this new bistro i \textcolor{blue}{discovered}... (A)\\ 
        5. \textcolor{ku}{lilly:} good idea &  
        7. \textcolor{jhu}{maya:} i'll get you 5 & \\ 
        6. \textcolor{ku}{lilly:} \colorbox{lime!}{pasta with salmon and basil} as always  &
        8. \textcolor{ku}{randolph:} thanks darling & \\ 
        very tasty there & & \\ \cline{3-3}
        \cline{1-2}
        {\bf REF:} lilly will be late. gabriel will order \colorbox{lime!}{pasta with salmon and basil} for her.&
        {\bf REF:} maya will buy 5 packs of earplugs for & \\
        & randolph at the pharmacy. & (2) Style C\\ \cline{1-2} \cline{1-2}
        & & {\shortstack[l]{...Iris : \textcolor{blue}{<file other>} My husband is famous...\\Haha. You don't even realize what this...}}\\
        {\shortstack[l]{{\bf FE:} lilly will be late. lilly and gabriel are going \\ to pasta with salmon and basil is always tasty.\textcolor{orange}{[62/46/68]}}} &
        {\shortstack[l]{{\bf FE:} randolph is in the pharmacy. \colorbox{magenta!}{randolph} will buy some earplugs \\ for randolph. maya will get 5.\textcolor{orange}{[64/38/71]}}} & ...Dan : \textcolor{blue}{<photo file>}...But its not working any more and it hurts \textcolor{blue}{:(}...\\ 
        {\shortstack[l]{{\bf B:} lilly will be late. \colorbox{pink!}{gabriel and lilly} will order food for \colorbox{pink!}{lilly and gabriel}.\\ \textcolor{orange}{[72/39/63]}}} &
        {\shortstack[l]{{\bf B:} \colorbox{magenta!}{laurie} is in the pharmacy. maya will buy 4 or 5 pairs of earphones\\ for him. \textcolor{orange}{[51/23/51]}}} &{\shortstack[l]{...Simon : \textcolor{blue}{BTW} it's so annoying that people can't see that \\such immigration policy reduces...}}\\ \cline{1-2}
        {\shortstack[l]{{\bf SB:} lilly is going to be late. gabriel will order food for her. \\ \colorbox{lime!}{lilly will get pasta with salmon and basil} and \\ \colorbox{cyan}{she will get it as soon as she arrives at them.}\textcolor{orange}{[68/23/68]}}} &
        {\shortstack[l]{{\bf SB:} maya will buy \colorbox{cyan!}{4} or 5 pairs of earplugs \\ for \colorbox{magenta!}{raymond} at the pharmacy. \textcolor{orange}{[63/34/63]}}} & \\ 
        \bottomrule
    \end{tabular}%
    }
    \caption{Examples of dialogues from each model. REF -- reference summary, FE -- Fast Abs RL Enhanced, B -- vanilla BART, and SB -- Syntax-aware BART (Ours). \textcolor{orange}{[R-1/R-2/R-L]} indicates F1 score from ROUGE-n. The error consists of the following factors \colorbox{pink!}{(i)}, \colorbox{magenta!}{(ii)}, and \colorbox{cyan}{(iii)}; otherwise, the accurate case is colored \colorbox{lime!}{lime}.}
    \label{tab:qualitative-1}
\end{table*}

\subsection{Qualitative Results}
We compared the generated examples from several baselines including our proposed model in terms of their ROUGE scores. We present the qualitative results in Table~\ref{tab:qualitative-1}. We observed the error analysis through the following major error types -- \textit{(i) Incorrect reasoning}: indicates that the model came to the incorrect conclusion, which occurred when the generated summaries reasoned relations in the dialogue incorrectly. \textit{(ii) Incorrect reference}: indicates the association of one's locations or actions with an incorrect speaker, regardless of the original context in the generated summaries. \textit{(iii) Redundancy}: is the case wherein the content of the generated summaries was not mentioned in a reference. \textit{(iv) Missing information}: content existing in the reference is absent in generated summaries.
\paragraph{Error anaylsis}
In dialogue 1, FE ({\em{i.e.,}} Fast Abs RL Enhanced) and SB ({\em{i.e.,}} Syntax-aware BART) performed well at capturing the meaning of the reference summary, despite it being slightly lengthy. The B model ({\em{i.e.,}}vanilla BART model) showed the highest score but contained instances of (i) incorrect reasoning (\textit{gabriel and lilly}). However, our SB model highlighted ({\em{i.e.,}} lime-colored) the content influenced by the lengthy utterance at line 6. It appears that the model was affected by the lengthy input type. Alternatively, there is also the (iii) Redundancy case, as shown in SB. The related content appeared in dialogue but was absent from the reference.
\begin{figure}[hbt!]
\begin{center}
\includegraphics[width=0.5\linewidth]{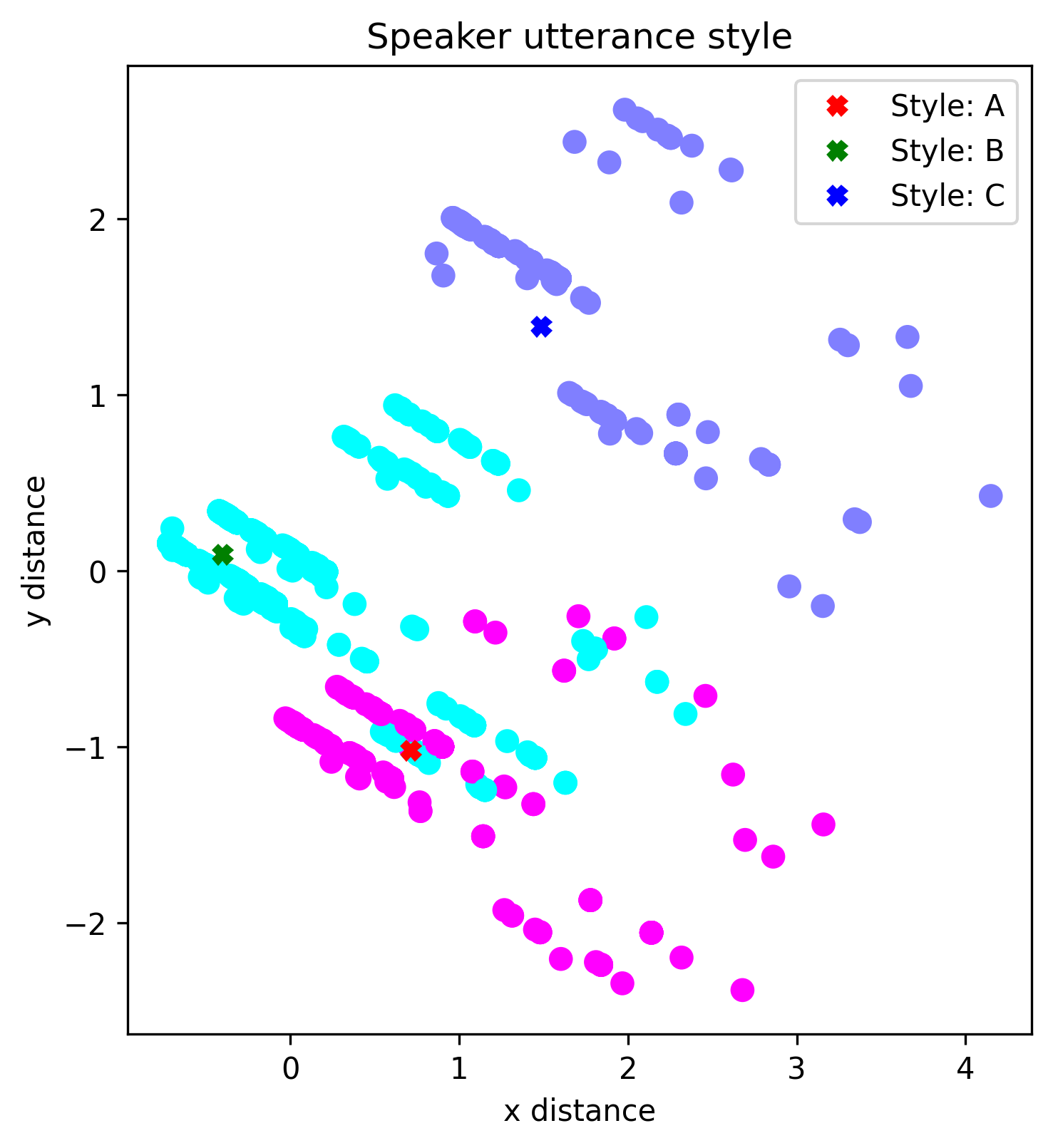}
\end{center}
\caption{Two-dimensional PCA projection of each speaker style - A (margenta), B (blue), and C (purple). The legend indicates the center point of each cluster. } 
\label{fig:speaker_pca}
\end{figure} 
\begin{figure}[hbt!]
\begin{center}
\includegraphics[width=0.8\linewidth]{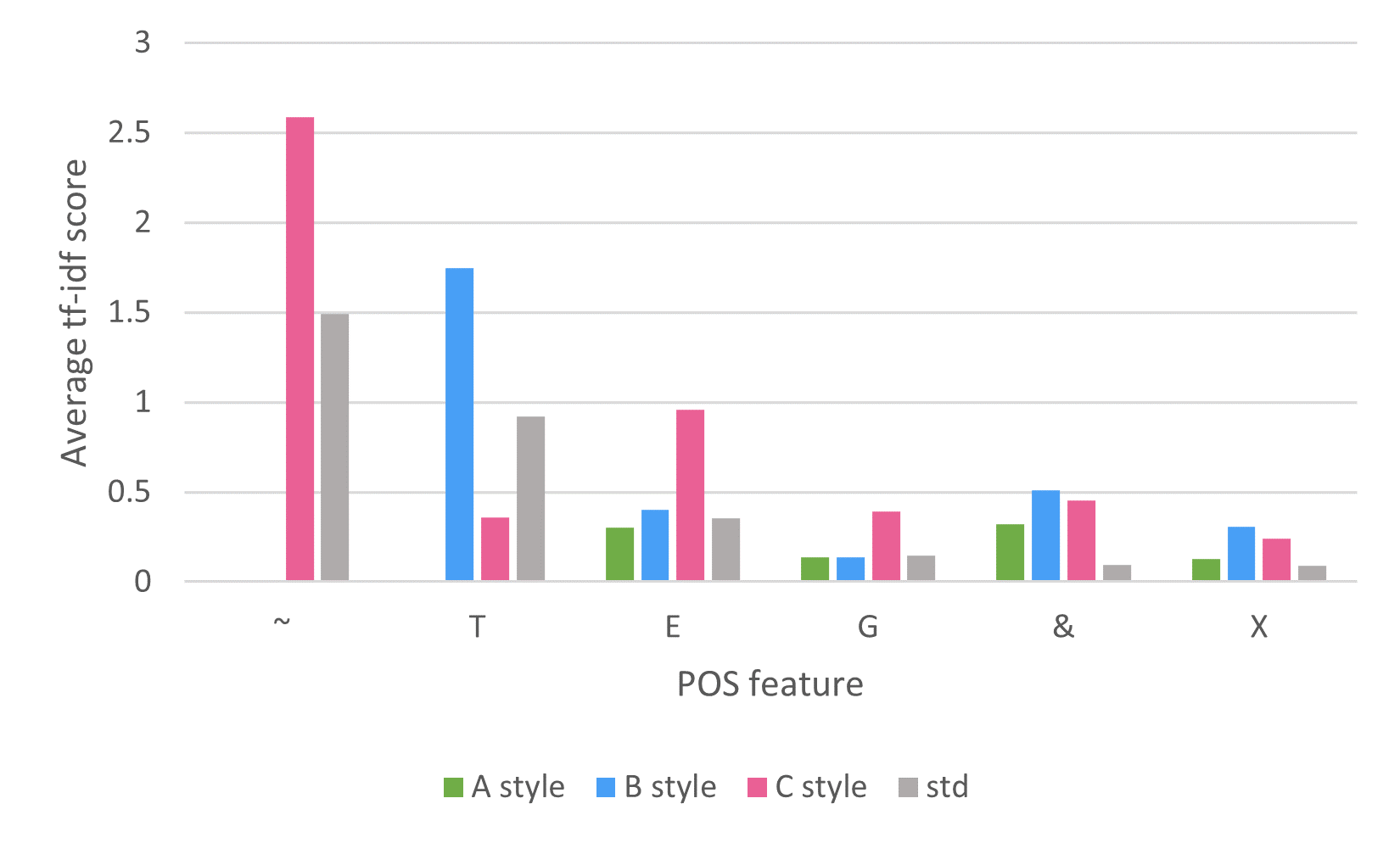}
\end{center}
\caption{Average tf-idf score on top six ranked POS features by standard deviation (std) according to the speaker styles (A, B, and C).} 
\label{fig:group_analysis}
\end{figure} 
In dialogue 2, the SB model showed the highest performance. FE mismatched (ii) the information of who was acting ({\em{i.e.,}} the subject) and who received the action ({\em{i.e.,}} the object). This observation also true of the B (\textit{laurie}) and SB (\textit{raymond}) models. Additionally, there was a case of (iii) redundancy in the SB (\textit{4}) model, despite existing content being present in the dialogue. 
In sum, according to the observations listed in Table~\ref{tab:qualitative-1}, our proposed model captured the lengthy utterance well in terms of our objective to use the locational information. Nonetheless, there are limitations to this study, such as the inclusion of incorrect references. This is an area that we aim to improve in future work.
\begin{table}[H]
\centering
\resizebox{\columnwidth}{!}{
\begin{tabular}{c|cccccc}
\hline
            & \textbf{$\sim$}           & \textbf{T}             & \textbf{E}        & \textbf{G}                                                                     & \textbf{\& }                                                                 & \textbf{X}                                                                                                      \\ \hline
Description & discourse/file marker & verb particle & emoticon & \begin{tabular}[c]{@{}c@{}}abbreviation,\\ foreign words\end{tabular} & \begin{tabular}[c]{@{}c@{}}coordinating \\ conjunction\end{tabular} & \begin{tabular}[c]{@{}c@{}}existential \textbackslash{}textit\{there\}, \\ predeterminers\end{tabular} \\ \hline
Example     & @user:hello    & up, out       & ;-), :b  & btw (by the way)                                                      & and, but                                                            & both, all, half                                                                                        \\ \hline
\end{tabular}}
\caption{POS tag description for the top-6 ranked.} 
\label{tab:tag_info}
\end{table}
\paragraph{Speaker utterance style (Vitamin)}
We discovered the speaker styles from the test set by following the Section~\ref{sec:speaker_style} for ad-hoc analysis. In details, our research question\footnote{We carried out the group characteristics in Appendix~\ref{appendix:speaker_style}.} was \textit{``what are the differences between the speakers?''}.
Figure~\ref{fig:speaker_pca} depicts the distributional characteristics through PCA (principal components analysis) projection for each speaker style and K-means clustering (K=3). The style `A' and `B' have some intersection what is even shown in Figure~\ref{fig:group_analysis}, and `C' is relatively distant from other groups.

In Figure~\ref{fig:group_analysis}, we illustrate the top-6 ranked POS features to distinguish the groups (see the detail values in Appendix~\ref{appendix:speaker_style}). Style `B' specifies \textup{T} that represent the verb particle, and style `C' mainly consists of \textup{$\sim$, E, G} as different factors than other groups. However, style `A' shows the relatively flatten performance below 1.0 as Figure~\ref{fig:group_analysis} and Appendix~\ref{appendix:speaker_style}. In Table~\ref{tab:qualitative-1}'s right table, we compared the speaker styles: (1) Style `B' used verb particle (\textit{find out}), but `A' used the same representation in a different way(\textit{discovered}) and (2) Style `C' mostly tend to represent their intention using picture or contents. In the end, we found that the ability to distinguish those speaker styles was reflected in our model.

\section{Conclusion}
In this study, we proposed a novel syntax-aware sequence-to-sequence model that leverages syntactic information (i.e., POS tagging), considering the informal daily chat structure constraints, and distinguishes the different textual styles from multiple speakers for abstractive dialogue summarization. To strategically combine syntactic information to the dialogue summarization task, we adopted multi-task learning to reproduce both syntactic information and dialogue summarization. Furthermore, we presented a novel input type to train the model to explore locational biases in dialogue structures.
We benchmarked the experiments using the SAMSum corpus, and the experimental results demonstrate that the proposed method improves comparison models for all ROUGE scores. 

There are promising future directions regarding this research. It would be worthwhile to apply the traditional truncation method with our proposed model to deeply compare performance differences. 



\bibliography{anthology,custom}

\begin{thebibliography}{40}
\expandafter\ifx\csname natexlab\endcsname\relax\def\natexlab#1{#1}\fi

\bibitem[{Afsharizadeh et~al.(2018)Afsharizadeh, Ebrahimpour-Komleh, and
  Bagheri}]{afsharizadeh2018query}
Mahsa Afsharizadeh, Hossein Ebrahimpour-Komleh, and Ayoub Bagheri. 2018.
\newblock Query-oriented text summarization using sentence extraction
  technique.
\newblock In \emph{2018 4th International Conference on Web Research (ICWR)},
  pages 128--132. IEEE.

\bibitem[{Al-sharman and Pivkina(2018)}]{al2018generating}
Nesreen Al-sharman and Inna~V Pivkina. 2018.
\newblock Generating summaries through selective part of speech tagging.
\newblock In \emph{Proceedings of the Fourth International Conference on
  Engineering \& MIS 2018}, pages 1--4.

\bibitem[{Arifin et~al.(2018)Arifin, Abdullah, Rosyadi, Ulumi, Wahib, and
  Sholikah}]{arifin2018sentence}
Agus~Zainal Arifin, Moch~Zawaruddin Abdullah, Ahmad~Wahyu Rosyadi, Desepta~Isna
  Ulumi, Aminul Wahib, and Rizka~Wakhidatus Sholikah. 2018.
\newblock Sentence extraction based on sentence distribution and part of speech
  tagging for multi-document summarization.
\newblock \emph{Telkomnika}, 16(2):843--851.

\bibitem[{Bouras and Tsogkas(2008)}]{bouras2008improving}
Christos Bouras and Vassilis Tsogkas. 2008.
\newblock Improving text summarization using noun retrieval techniques.
\newblock In \emph{International Conference on Knowledge-Based and Intelligent
  Information and Engineering Systems}, pages 593--600. Springer.

\bibitem[{Cachola et~al.(2020)Cachola, Lo, Cohan, and
  Weld}]{cachola-etal-2020-tldr}
Isabel Cachola, Kyle Lo, Arman Cohan, and Daniel Weld. 2020.
\newblock \href {https://doi.org/10.18653/v1/2020.findings-emnlp.428} {{TLDR}:
  Extreme summarization of scientific documents}.
\newblock In \emph{Findings of the Association for Computational Linguistics:
  EMNLP 2020}, pages 4766--4777, Online. Association for Computational
  Linguistics.

\bibitem[{Chen and Yang(2020)}]{chen2020multi}
Jiaao Chen and Diyi Yang. 2020.
\newblock Multi-view sequence-to-sequence models with conversational structure
  for abstractive dialogue summarization.
\newblock \emph{arXiv preprint arXiv:2010.01672}.

\bibitem[{Chen and Bansal(2018)}]{chen2018fast}
Yen-Chun Chen and Mohit Bansal. 2018.
\newblock Fast abstractive summarization with reinforce-selected sentence
  rewriting.
\newblock \emph{arXiv preprint arXiv:1805.11080}.

\bibitem[{Dobreva et~al.(2020)Dobreva, Jofche, Jovanovik, and
  Trajanov}]{dobreva2020improving}
Jovana Dobreva, Nasi Jofche, Milos Jovanovik, and Dimitar Trajanov. 2020.
\newblock Improving ner performance by applying text summarization on
  pharmaceutical articles.
\newblock In \emph{International Conference on ICT Innovations}, pages 87--97.
  Springer.

\bibitem[{Erera et~al.(2019)Erera, Shmueli-Scheuer, Feigenblat, Peled~Nakash,
  Boni, Roitman, Cohen, Weiner, Mass, Rivlin, Lev, Jerbi, Herzig, Hou, Jochim,
  Gleize, Bonin, Bonin, and Konopnicki}]{erera-etal-2019-summarization}
Shai Erera, Michal Shmueli-Scheuer, Guy Feigenblat, Ora Peled~Nakash, Odellia
  Boni, Haggai Roitman, Doron Cohen, Bar Weiner, Yosi Mass, Or~Rivlin, Guy Lev,
  Achiya Jerbi, Jonathan Herzig, Yufang Hou, Charles Jochim, Martin Gleize,
  Francesca Bonin, Francesca Bonin, and David Konopnicki. 2019.
\newblock \href {https://doi.org/10.18653/v1/D19-3036} {A summarization system
  for scientific documents}.
\newblock In \emph{Proceedings of the 2019 Conference on Empirical Methods in
  Natural Language Processing and the 9th International Joint Conference on
  Natural Language Processing (EMNLP-IJCNLP): System Demonstrations}, pages
  211--216, Hong Kong, China. Association for Computational Linguistics.

\bibitem[{Gehrmann et~al.(2018)Gehrmann, Deng, and
  Rush}]{gehrmann-etal-2018-bottom}
Sebastian Gehrmann, Yuntian Deng, and Alexander Rush. 2018.
\newblock \href {https://doi.org/10.18653/v1/D18-1443} {Bottom-up abstractive
  summarization}.
\newblock In \emph{Proceedings of the 2018 Conference on Empirical Methods in
  Natural Language Processing}, pages 4098--4109, Brussels, Belgium.
  Association for Computational Linguistics.

\bibitem[{Gimpel et~al.(2010)Gimpel, Schneider, O'Connor, Das, Mills,
  Eisenstein, Heilman, Yogatama, Flanigan, and Smith}]{gimpel2010part}
Kevin Gimpel, Nathan Schneider, Brendan O'Connor, Dipanjan Das, Daniel Mills,
  Jacob Eisenstein, Michael Heilman, Dani Yogatama, Jeffrey Flanigan, and
  Noah~A Smith. 2010.
\newblock Part-of-speech tagging for twitter: Annotation, features, and
  experiments.
\newblock Technical report, Carnegie-Mellon Univ Pittsburgh Pa School of
  Computer Science.

\bibitem[{Gliwa et~al.(2019)Gliwa, Mochol, Biesek, and Wawer}]{gliwa2019samsum}
Bogdan Gliwa, Iwona Mochol, Maciej Biesek, and Aleksander Wawer. 2019.
\newblock Samsum corpus: A human-annotated dialogue dataset for abstractive
  summarization.
\newblock \emph{arXiv preprint arXiv:1911.12237}.

\bibitem[{Goo and Chen(2018)}]{goo2018abstractive}
Chih-Wen Goo and Yun-Nung Chen. 2018.
\newblock Abstractive dialogue summarization with sentence-gated modeling
  optimized by dialogue acts.
\newblock In \emph{2018 IEEE Spoken Language Technology Workshop (SLT)}, pages
  735--742. IEEE.

\bibitem[{Guo et~al.(2021)Guo, Yang, and Gao}]{guo2021speaker}
Miao Guo, Jiaxiong Yang, and Shu Gao. 2021.
\newblock Speaker recognition method for short utterance.
\newblock In \emph{Journal of Physics: Conference Series}, volume 1827, page
  012158. IOP Publishing.

\bibitem[{Jones et~al.(2004)Jones, Ravid, and Rafaeli}]{jones2004information}
Quentin Jones, Gilad Ravid, and Sheizaf Rafaeli. 2004.
\newblock Information overload and the message dynamics of online interaction
  spaces: A theoretical model and empirical exploration.
\newblock \emph{Information systems research}, 15(2):194--210.

\bibitem[{Kim et~al.(2019)Kim, Kim, and Kim}]{kim2019abstractive}
Byeongchang Kim, Hyunwoo Kim, and Gunhee Kim. 2019.
\newblock Abstractive summarization of reddit posts with multi-level memory
  networks.
\newblock In \emph{NAACL-HLT (1)}.

\bibitem[{Kingma and Ba(2014)}]{kingma2014adam}
Diederik~P Kingma and Jimmy Ba. 2014.
\newblock Adam: A method for stochastic optimization.
\newblock \emph{arXiv preprint arXiv:1412.6980}.

\bibitem[{Koppel et~al.(2009)Koppel, Schler, and
  Argamon}]{koppel2009computational}
Moshe Koppel, Jonathan Schler, and Shlomo Argamon. 2009.
\newblock Computational methods in authorship attribution.
\newblock \emph{Journal of the American Society for information Science and
  Technology}, 60(1):9--26.

\bibitem[{K{\"u}bler et~al.(2010)K{\"u}bler, Scheutz, Baucom, and
  Israel}]{kubler2010adding}
Sandra K{\"u}bler, Matthias Scheutz, Eric Baucom, and Ross Israel. 2010.
\newblock Adding context information to part of speech tagging for dialogues.
\newblock In \emph{Ninth International Workshop on Treebanks and Linguistic
  Theories}, page 115.

\bibitem[{Lewis et~al.(2020)Lewis, Liu, Goyal, Ghazvininejad, Mohamed, Levy,
  Stoyanov, and Zettlemoyer}]{lewis-etal-2020-bart}
Mike Lewis, Yinhan Liu, Naman Goyal, Marjan Ghazvininejad, Abdelrahman Mohamed,
  Omer Levy, Veselin Stoyanov, and Luke Zettlemoyer. 2020.
\newblock \href {https://doi.org/10.18653/v1/2020.acl-main.703} {{BART}:
  Denoising sequence-to-sequence pre-training for natural language generation,
  translation, and comprehension}.
\newblock In \emph{Proceedings of the 58th Annual Meeting of the Association
  for Computational Linguistics}, pages 7871--7880, Online. Association for
  Computational Linguistics.

\bibitem[{Lin(2004)}]{lin2004rouge}
Chin-Yew Lin. 2004.
\newblock Rouge: A package for automatic evaluation of summaries.
\newblock In \emph{Text summarization branches out}, pages 74--81.

\bibitem[{Liu and Wang(2017)}]{liu2017pos}
Wuying Liu and Lin Wang. 2017.
\newblock Pos-tagging enhanced korean text summarization.
\newblock In \emph{International Conference on Intelligent Computing}, pages
  425--435. Springer.

\bibitem[{Liu et~al.(2018)Liu, Wu, Li, Li, and Shen}]{liu2018gmm}
Zheli Liu, Zhendong Wu, Tong Li, Jin Li, and Chao Shen. 2018.
\newblock Gmm and cnn hybrid method for short utterance speaker recognition.
\newblock \emph{IEEE Transactions on Industrial informatics}, 14(7):3244--3252.

\bibitem[{Lu et~al.(2019)Lu, Hou, Wang, Huang, Fei, and
  Zhang}]{lu2019attributed}
Ruqian Lu, Shengluan Hou, Chuanqing Wang, Yu~Huang, Chaoqun Fei, and Songmao
  Zhang. 2019.
\newblock Attributed rhetorical structure grammar for domain text
  summarization.
\newblock \emph{arXiv preprint arXiv:1909.00923}.

\bibitem[{McCowan et~al.(2005)McCowan, Carletta, Kraaij, Ashby, Bourban, Flynn,
  Guillemot, Hain, Kadlec, Karaiskos et~al.}]{mccowan2005ami}
Iain McCowan, Jean Carletta, Wessel Kraaij, Simone Ashby, S~Bourban, M~Flynn,
  M~Guillemot, Thomas Hain, J~Kadlec, Vasilis Karaiskos, et~al. 2005.
\newblock The ami meeting corpus.
\newblock In \emph{Proceedings of the 5th International Conference on Methods
  and Techniques in Behavioral Research}, volume~88, page 100. Citeseer.

\bibitem[{Nallapati et~al.(2016)Nallapati, Zhou, Gulcehre, Xiang
  et~al.}]{nallapati2016abstractive}
Ramesh Nallapati, Bowen Zhou, Caglar Gulcehre, Bing Xiang, et~al. 2016.
\newblock Abstractive text summarization using sequence-to-sequence rnns and
  beyond.
\newblock \emph{arXiv preprint arXiv:1602.06023}.

\bibitem[{Owoputi et~al.(2013)Owoputi, O’Connor, Dyer, Gimpel, Schneider, and
  Smith}]{owoputi2013improved}
Olutobi Owoputi, Brendan O’Connor, Chris Dyer, Kevin Gimpel, Nathan
  Schneider, and Noah~A Smith. 2013.
\newblock Improved part-of-speech tagging for online conversational text with
  word clusters.
\newblock In \emph{Proceedings of the 2013 conference of the North American
  chapter of the association for computational linguistics: human language
  technologies}, pages 380--390.

\bibitem[{Pal and Saha(2014)}]{pal2014approach}
Alok~Ranjan Pal and Diganta Saha. 2014.
\newblock An approach to automatic text summarization using wordnet.
\newblock In \emph{2014 IEEE International Advance Computing Conference
  (IACC)}, pages 1169--1173. IEEE.

\bibitem[{Paulus et~al.(2017)Paulus, Xiong, and Socher}]{paulus2017deep}
Romain Paulus, Caiming Xiong, and Richard Socher. 2017.
\newblock A deep reinforced model for abstractive summarization.
\newblock \emph{arXiv preprint arXiv:1705.04304}.

\bibitem[{Precht(2008)}]{precht2008sex}
Kristen Precht. 2008.
\newblock Sex similarities and differences in stance in informal american
  conversation 1.
\newblock \emph{Journal of Sociolinguistics}, 12(1):89--111.

\bibitem[{Radford et~al.(2019)Radford, Wu, Child, Luan, Amodei, and
  Sutskever}]{radford2019language}
Alec Radford, Jeffrey Wu, Rewon Child, David Luan, Dario Amodei, and Ilya
  Sutskever. 2019.
\newblock Language models are unsupervised multitask learners.
\newblock \emph{OpenAI blog}, 1(8):9.

\bibitem[{Sacks et~al.(1978)Sacks, Schegloff, and
  Jefferson}]{sacks1978simplest}
Harvey Sacks, Emanuel~A Schegloff, and Gail Jefferson. 1978.
\newblock A simplest systematics for the organization of turn taking for
  conversation.
\newblock In \emph{Studies in the organization of conversational interaction},
  pages 7--55. Elsevier.

\bibitem[{See et~al.(2017)See, Liu, and Manning}]{see2017get}
Abigail See, Peter~J Liu, and Christopher~D Manning. 2017.
\newblock Get to the point: Summarization with pointer-generator networks.
\newblock In \emph{Proceedings of the 55th Annual Meeting of the Association
  for Computational Linguistics (Volume 1: Long Papers)}, pages 1073--1083.

\bibitem[{Shi et~al.(2021)Shi, Keneshloo, Ramakrishnan, and
  Reddy}]{shi2021neural}
Tian Shi, Yaser Keneshloo, Naren Ramakrishnan, and Chandan~K Reddy. 2021.
\newblock Neural abstractive text summarization with sequence-to-sequence
  models.
\newblock \emph{ACM Transactions on Data Science}, 2(1):1--37.

\bibitem[{Vinyals et~al.(2015)Vinyals, Fortunato, and
  Jaitly}]{vinyals2015pointer}
Oriol Vinyals, Meire Fortunato, and Navdeep Jaitly. 2015.
\newblock Pointer networks.
\newblock \emph{arXiv preprint arXiv:1506.03134}.

\bibitem[{Wu et~al.(2019)Wu, Fan, Baevski, Dauphin, and Auli}]{wu2019pay}
Felix Wu, Angela Fan, Alexei Baevski, Yann~N Dauphin, and Michael Auli. 2019.
\newblock Pay less attention with lightweight and dynamic convolutions.
\newblock \emph{arXiv preprint arXiv:1901.10430}.

\bibitem[{Yuan and Yu(2019)}]{yuan2019abstractive}
Lin Yuan and Zhou Yu. 2019.
\newblock Abstractive dialog summarization with semantic scaffolds.
\newblock \emph{arXiv preprint arXiv:1910.00825}.

\bibitem[{Zhang et~al.(2019)Zhang, Kishore, Wu, Weinberger, and
  Artzi}]{zhang2019bertscore}
Tianyi Zhang, Varsha Kishore, Felix Wu, Kilian~Q Weinberger, and Yoav Artzi.
  2019.
\newblock Bertscore: Evaluating text generation with bert.
\newblock \emph{arXiv preprint arXiv:1904.09675}.

\bibitem[{Zhu et~al.(2020)Zhu, Xu, Zeng, and Huang}]{zhu2020hierarchical}
Chenguang Zhu, Ruochen Xu, Michael Zeng, and Xuedong Huang. 2020.
\newblock A hierarchical network for abstractive meeting summarization with
  cross-domain pretraining.
\newblock \emph{arXiv preprint arXiv:2004.02016}.

\bibitem[{Zopf et~al.(2018)Zopf, Botschen, Falke, Heinzerling, Marasovic,
  Mihaylov, Avinesh, Menc{\'\i}a, F{\"u}rnkranz, and Frank}]{zopf2018s}
Markus Zopf, Teresa Botschen, Tobias Falke, Beniamin Heinzerling, Ana
  Marasovic, Todor Mihaylov, PVS Avinesh, Eneldo~Loza Menc{\'\i}a, Johannes
  F{\"u}rnkranz, and Anette Frank. 2018.
\newblock What's important in a text? an extensive evaluation of linguistic
  annotations for summarization.
\newblock In \emph{2018 Fifth International Conference on Social Networks
  Analysis, Management and Security (SNAMS)}, pages 272--277. IEEE.

\end{thebibliography}
\bibliographystyle{acl_natbib}

\clearpage
\appendix

\section{Related Work} \label{appendix:related_work}
\subsection{Dialogue summarization}
Most of the previous works focused on summarizing conversations from meetings~\cite{mccowan2005ami}, in part due to the lack of a corpus for daily conversation summaries. Hence, SAMSum, which is a corpus of daily conversation summaries, has been receiving much attention as announced by~\citet{gliwa2019samsum} recently. 

In the standard dialogue summarization paradigm, a pointer-generator~\cite{see2017get}, which is a hybrid model of the typical sequence-to-sequence attention model~\cite{nallapati2016abstractive}, and a pointer network~\cite{vinyals2015pointer} are used as the abstractive dialogue summarization model. 
This framework encodes the source sequence and generates the target sequence, with the decoder for abstractive dialogue summarization ~\cite{yuan2019abstractive,goo2018abstractive}. 

Recently, the standard paradigm has shifted to using a combination of a pretraining method with much larger external text corpus ({\em{e.g.,}} Wikipedia, books) and a transformer-based sequence model. This strategy has led to a remarkable improvement in performance when fine-tuned for both text generation tasks and natural language understanding like BART~\cite{lewis-etal-2020-bart}. 


\subsection{Syntax-aware text summarization}
Syntax representation of text can be applied to text summarization to leverage linguistic information because it assists in information filtering to obtain highlighted context from a source document~\cite{bouras2008improving}, and yet the importance of this syntax has been previously underestimated~\cite{zopf2018s}. When linguistic information is used to perform text summarization, it finds the relationships between terms in the document through sequence labeling (POS tagging~\cite{al2018generating}, named entity recognition~\cite{dobreva2020improving}), grammar analysis~\cite{lu2019attributed}, and thesaurus usage ({\em{e.g.,}} Wordnet)~\cite{pal2014approach}, and then extracts the salient context. 

Previous research has investigated the use of linguistic information such as POS tagging for text summarization. \citet{al2018generating} approached selective POS tagging for words such as nouns, verbs, and adjectives to extract sentence summaries. \citet{bouras2008improving} and \citet{afsharizadeh2018query} also attempted to extract keywords by retrieving nouns through POS tagging. \citet{liu2017pos} reported the utilization of POS tagging to distil keywords for extractive summarization in Korean.

However, these studies only considered extractive summarization tasks and applied linguistic information to extract the key sentences as features using the scoring function. By contrast, we propose to have a pretrained model learn linguistic information implicitly through the sequence labeling task to function in an abstractive way. 

\section{Speaker Style Cluster Result}\label{appendix:speaker_style}
Table~\ref{tab:speaker_cluster} shows POS features~\cite{gimpel2010part} by speaker styles. 
We calculated the average of tf-idf values according to each groups. Note that the extremely common terms occur tf-idf value to zero. The standard deviation shows how much there is the between-group deviation for each POS feature. 
\begin{table}[h!]
\centering
\scalebox{0.8}{
\begin{tabular}{c|l|l|l|l}
\toprule
\multicolumn{1}{l|}{feature} & A style & B style & C style & std  \\ \hline
$\sim$                        & 0       & 0       & 2.58    & 1.49 \\ \hline
T                             & 0       & 1.75    & 0.36    & 0.92 \\ \hline
E                             & 0.3     & 0.4     & 0.96    & 0.35 \\ \hline
G                             & 0.14    & 0.14    & 0.39    & 0.15 \\ \hline
\&                            & 0.32    & 0.51    & 0.45    & 0.1  \\ \hline
X                             & 0.13    & 0.31    & 0.24    & 0.09 \\ \hline
\$                            & 0.32    & 0.49    & 0.44    & 0.09 \\ \hline
Z                             & 0.09    & 0.18    & 0.22    & 0.07 \\ \hline
L                             & 0.29    & 0.4     & 0.33    & 0.06 \\ \hline
!                             & 0.27    & 0.37    & 0.31    & 0.05 \\ \hline
U                             & 0       & 0       & 0.05    & 0.03 \\ \hline
S                             & 0.06    & 0.09    & 0.06    & 0.02 \\ \hline
D                             & 0.19    & 0.22    & 0.22    & 0.02 \\ \hline
R                             & 0.19    & 0.22    & 0.2     & 0.02 \\ \hline
A                             & 0.19    & 0.23    & 0.21    & 0.02 \\ \hline
P                             & 0.17    & 0.2     & 0.18    & 0.02 \\ \hline
@                             & 0.01    & 0.02    & 0       & 0.01 \\ \hline
O                             & 0.1     & 0.12    & 0.11    & 0.01 \\ \hline
N                             & 0.09    & 0.1     & 0.1     & 0.01 \\ \hline
V                             & 0.07    & 0.08    & 0.07    & 0    \\ \hline
Y                             & 0       & 0       & 0       & 0    \\ \hline
\textasciicircum{}            & 0       & 0       & 0       & 0    \\ \hline
,                             & 0       & 0       & 0       & 0    \\ 
\bottomrule
\end{tabular}}
\caption{Average tf-idf score of each speaker styles. Ordered by standard deviation (std). A-\img{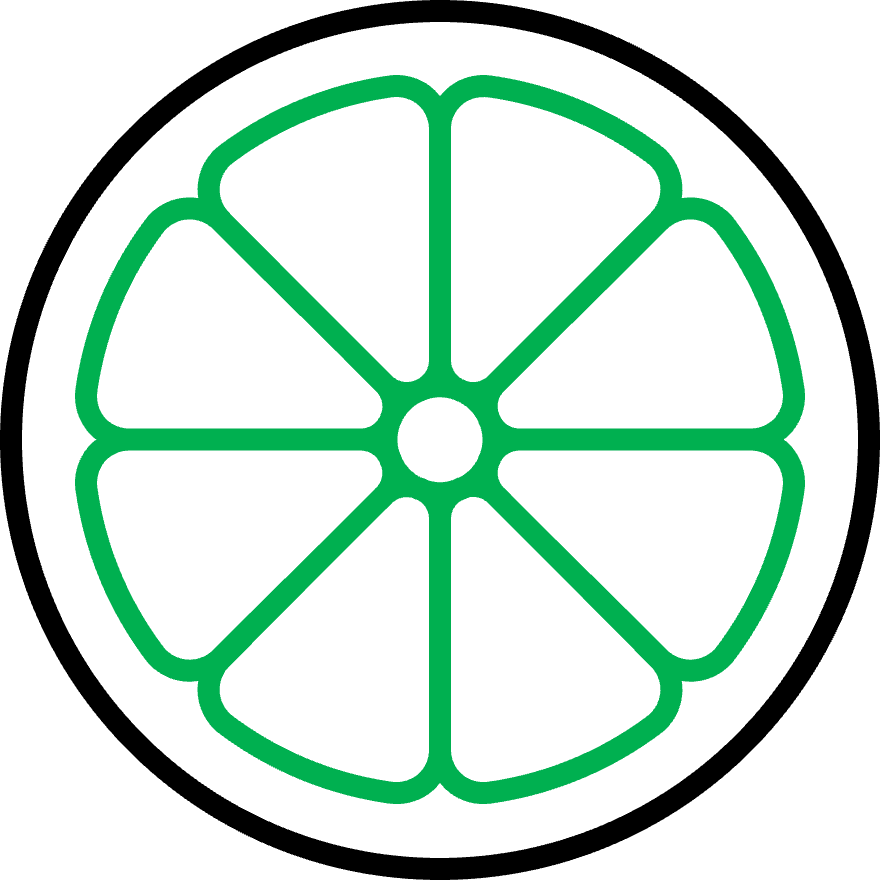}, B-\img{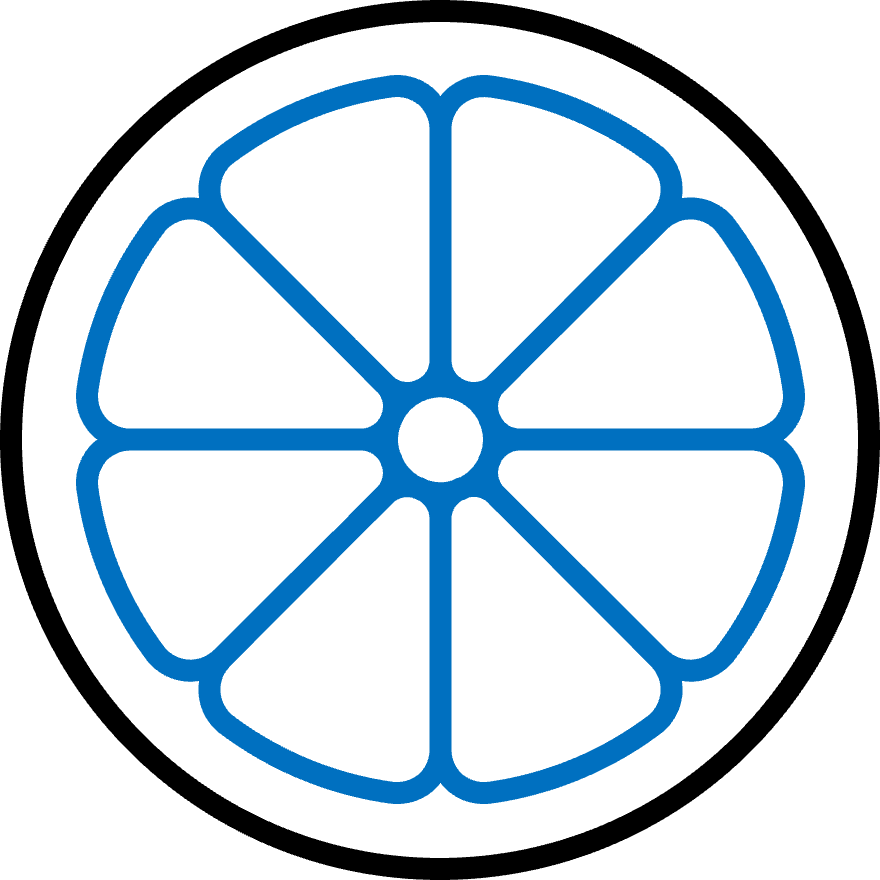}, and C-\img{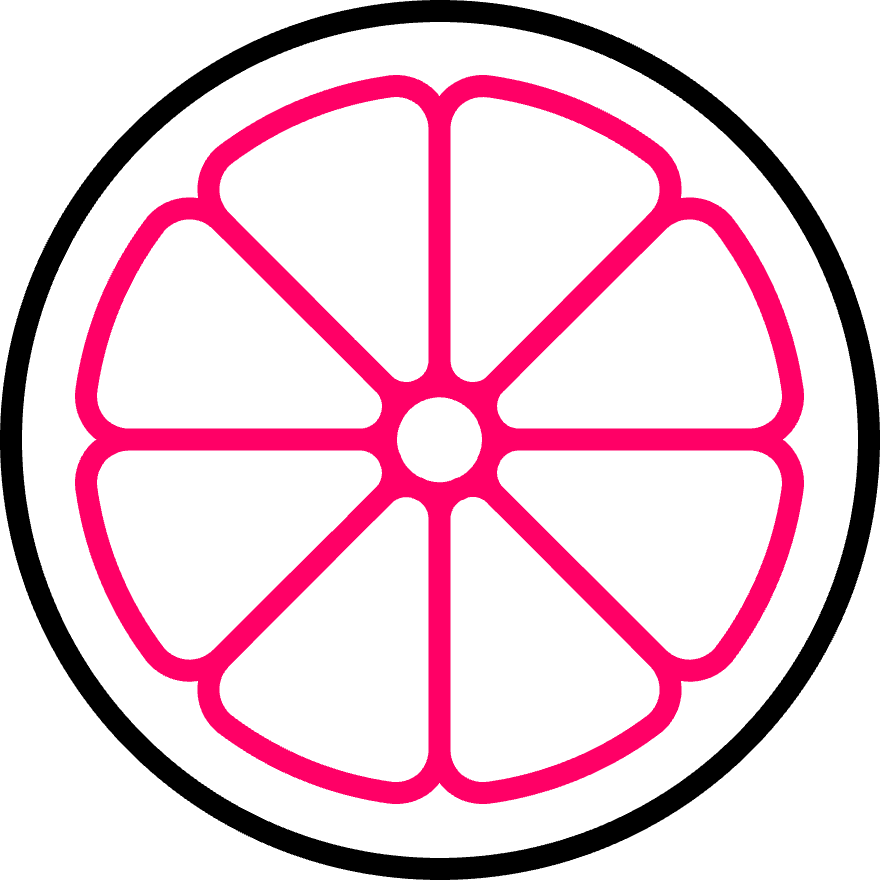}.} \label{tab:speaker_cluster}
\end{table}

\section{Density of the number of utterances}\label{appendix:density}
\begin{figure}[H] 
\begin{center}
\includegraphics[width=0.8\linewidth]{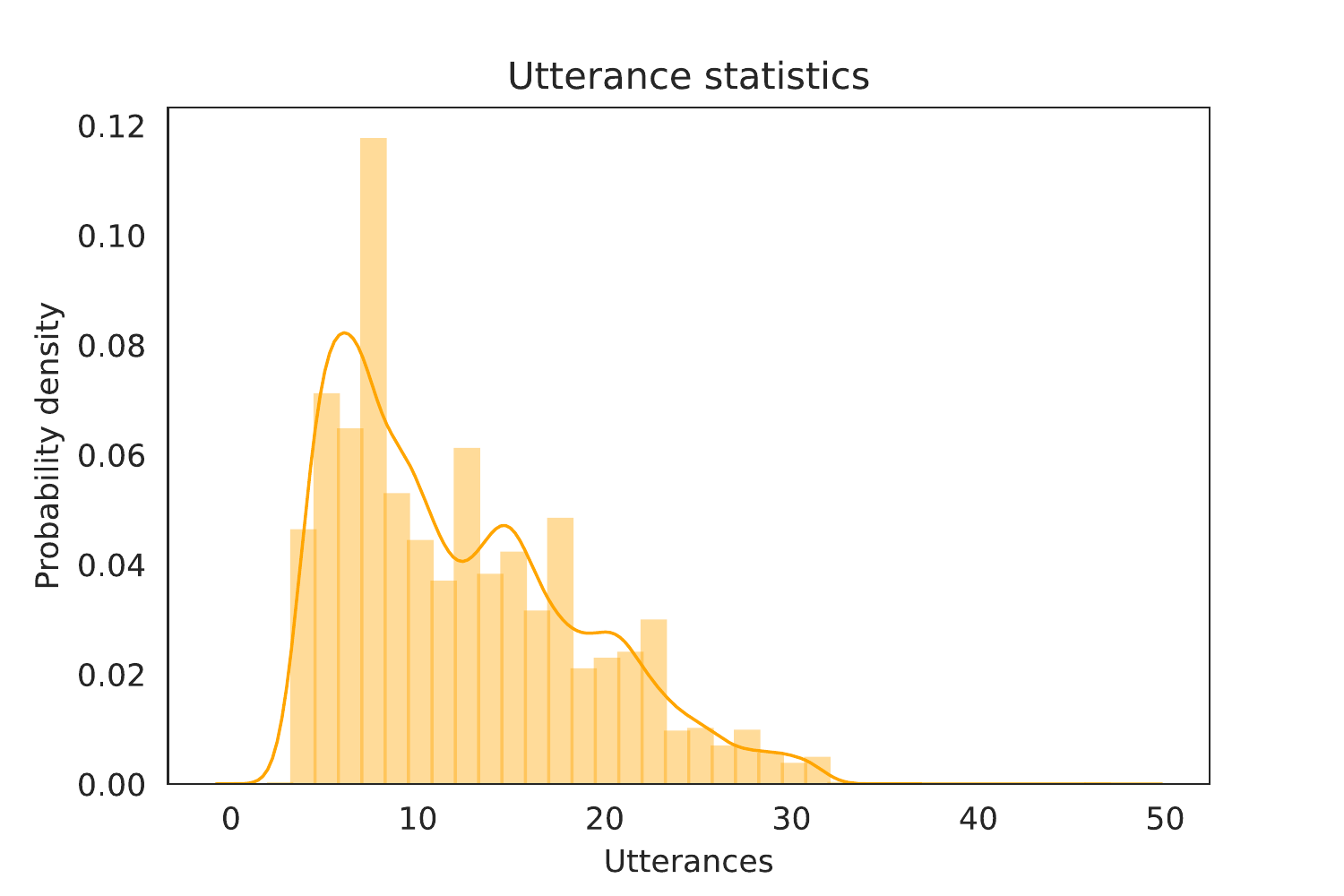}
\end{center}
\caption{Data distribution of the number of utterances in the SAMSum corpus (training set).} 
\label{fig:utter-statistic}
\end{figure} 

\end{document}